# The $G_{n,m}$ Phase Transition is Not Hard for the Hamiltonian Cycle Problem

**Basil Vandegriend**                                    BASIL@CS.UALBERTA.CA
**Joseph Culberson**                                     JOE@CS.UALBERTA.CA
*Department of Computing Science, University of Alberta,*
*Edmonton, Alberta, Canada, T6G 2H1*

## Abstract

Using an improved backtrack algorithm with sophisticated pruning techniques, we revise previous observations correlating a high frequency of hard to solve Hamiltonian cycle instances with the $G_{n,m}$ phase transition between Hamiltonicity and non-Hamiltonicity. Instead all tested graphs of 100 to 1500 vertices are easily solved.

When we artificially restrict the degree sequence with a bounded maximum degree, although there is some increase in difficulty, the frequency of hard graphs is still low. When we consider more regular graphs based on a generalization of knight's tours, we observe frequent instances of really hard graphs, but on these the average degree is bounded by a constant. We design a set of graphs with a feature our algorithm is unable to detect and so are very hard for our algorithm, but in these we can vary the average degree from $O(1)$ to $O(n)$. We have so far found no class of graphs correlated with the $G_{n,m}$ phase transition which asymptotically produces a high frequency of hard instances.

## 1. Introduction

Given a graph $G = (V, E), |V| = n, |E| = m$, the Hamiltonian cycle problem is to find a cycle $C = (v_1, v_2, \ldots, v_n)$ such that $v_i \neq v_j$ for $i \neq j$, $(v_i, v_{i+1}) \in E$ and $(v_n, v_1) \in E$. As for any NP-C problem, we expect solving it to require exponential time in the worst case on arbitrary graphs (assuming P $\neq$ NP). However, in recent years researchers examining various NP-C problems such as SAT and graph coloring have discovered that the majority of graphs are easy for their algorithms to solve. Only graphs with specific characteristics or graphs which lie within a narrow band (according to some parameter) seem to be hard to solve for these problems.

It is known (Pósa, 1976; Komlós & Szemerédi, 1983) that under a random graph model ($G_{n,m}$) as the edge density increases there is a sharp threshold (the phase transition) such that below that edge density the probability of a Hamiltonian cycle is 0, while above it the probability is 1. Previous research (Section 2.1) suggested that there is a high correlation of difficult problems with instances generated with edge density near the phase transition. Using an improved Hamiltonian cycle backtrack algorithm (Section 3) that employs various pruning operators and an iterated restart technique, we observe no hard instances at the transition for large $n$. Section 4 describes our results on $G_{n,m}$ and related random graphs.

In an attempt to find a higher frequency of hard graphs, in Section 5 we examine a low degree random graph class we call Degreebound graphs. However, these graphs are also usually easy for our backtrack algorithm, although we do find a few hard graphs. Analysis of these graphs indicates a test for non-Hamiltonian instances discussed in Section 5.3. In Section 6 we examine a graph class based on a generalization of the knight's tour problem. These graphs are significantly harder for our algorithm in general. In Section 7 we present a constructed graph class which produces exponential behavior for our backtrack algorithm.

Our experimental results provide evidence that the average degree of a graph is not a sufficient indicator for hard graphs for the Hamiltonian cycle problem. With our backtrack algorithm, the phase transition regions of the $G_{n,m}$ and Degreebound graph models are generally asymptotically easy.

## 2. A Discussion of Hardness and Previous Work

The concept of hardness of instances and hard regions within graph classes, considered from an empirical basis, is not easy to define. In order to clarify what we mean, in this section we present our notions of hardness, relating this to previous work.

### 2.1 What is Hardness?

A *problem of size n* is a set $\Pi_n$ of *instances*. For the Hamiltonian cycle problem, $\Pi_n$ is the set of undirected graphs on $n$ vertices. Any discussion of the hardness of a particular instance of a problem is always with respect to an algorithm (or set of algorithms). In general, different algorithms will perform differently on the instance. Furthermore, for each particular instance of Hamiltonian cycle there is an associated algorithm that either correctly answers NO or outputs a cycle in $O(n)$ time. To meaningfully talk about the hardness of an instance, we must assume a fixed algorithm (or a finite class of algorithms) that is appropriate for a large (infinite) class of instances, and then consider how the algorithm performs on the instance. Hardness of an instance is always a measure of performance relative to an algorithm.

We are left with the question of how much work an algorithm must do before we consider the instance hard for it. Note that for a *single* instance the distinction between polynomial and exponential time is moot. Ideally, we would like to require the algorithm to take an exponential (i.e. $a^n$ for some $a > 1$) number of steps as size $n$ increases. Note that empirical corroboration of such is practically impossible for sets of large instances. In practice, we must be content with evidence such as failure to complete within a reasonable time for larger instances.

We would also like an instance to exhibit some robustness before we consider it hard for a given algorithm. Ideally, for graph problems we would at a minimum require the instance to remain hard with high probability under a random relabeling of the vertices. Relabeling the vertices produces an isomorphic copy of the graph, preserving structural properties such as degree, connectivity, Hamiltonicity, cut sets, etc. The design of algorithms is typically based on identifying and using such properties, and as far as possible efficiency should be independent of the arbitrary assignment of labels.

Let us refer to a (probabilistic) *problem class* as a pair $(\Pi_n, \mathcal{P}_n)$, where $\mathcal{P}_n(x)$ is the probability of the instance $x$ given that we are selecting from $\Pi_n$. Problem classes are sometimes called *ensembles* in the Artificial Intelligence literature (Hogg, 1998). The usual classes for graph problems are $G_{n,p}$, where to generate an $n$ vertex graph, each pair of vertices is included as an edge with probability $p$, and $G_{n,m}$ where $m$ distinct edges are selected at random and placed in the graph. These two models are related (Palmer 1985). For this paper we use the $G_{n,m}$ model.

We do not consider mean or average run times in our definitions. The primary reason is that for exponentially small sets of exponentially hard instances, it is impractical to determine the average with any reasonable assurance. For example, if $1/2^n$ of the instances require $\Theta(n^2 2^n)$ time and the remainder are solved in $O(n^2)$ time then the average time is quadratic, while if the frequency increases to $1/2^{0.9n}$ the average time is exponential. Even for $n = 100$ it would be utterly impractical to distinguish between these two frequencies with empirical studies.

Furthermore, and for similar reasons, if we want to promote a class as a benchmark class for testing and comparing algorithms, low frequencies of hard instances are not generally sufficient. We will say that a problem class is *maximally hard* (with respect to an algorithm or set of algorithms) if the instances generated according to the distribution are hard with probability going to one as $n$ goes to infinity.

As an example of maximally hard classes, empirical evidence suggests that a variety of hidden coloring graph generators based on the $G_{n,p}$ model are maximally hard for a large variety of graph coloring algorithms (Culberson and Luo, 1993). These hard classes are all closely related to a coloring phase transition in random graphs. In general, a phase transition is defined by some parameterized probability distribution on the set of instances. As the parameter is varied past a certain threshold value, the asymptotic probability of the existence of a solution switches sharply from zero to one.

Phase transitions are commonly considered to be identified with hard subsets of a particular problem (Cheeseman, Kanefsky, & Taylor, 1991). Many NP-C problems can be characterized by a 'constraint' parameter which measures how constrained an instance is. Evaluation of a problem using this constraint parameter typically divides instances into two classes: those that are solvable, and those that are unsolvable, with a sharp transition occurring between them. When the problem is highly constrained, it is easily determined that no solution exists. As constraints are removed, a solution is easily found.

Different researchers (Cheeseman et al., 1991; Frank & Martel, 1995; Frank, Gent, & Walsh, 1998) have examined phase transitions on random graphs for the Hamiltonian cycle problem. The obvious constraint parameter is the average degree (or average connectivity) of the graph. As the degree increases, the graph becomes less constrained: it becomes easier both for a Hamiltonian cycle to exist and for an algorithm to find one. These researchers have examined how Hamiltonicity changes with respect to the average degree. Frank et al. (1998) and Frank and Martel (1995) experimentally verified that when using the $G_{n,m}$ model the phase transition for Hamiltonicity is very close to the phase transition for biconnectivity, which occurs when the average degree is approximately $\ln n$ (or $m = n \ln n / 2$) [1]. Cheeseman et al. (1991) experimentally confirmed theoretical predictions by Komlós and Szemerédi (1983) that the phase transition (for the Hamiltonian cycle problem) occurs when the average degree is $\ln n + \ln \ln n$. The papers also provided empirical evidence that the time required by their backtrack algorithms increased in the region of the phase transition and noted that the existence of very hard instances appeared to be associated with this transition.

---

1. Note that the average degree equals $2m/n$.

As mentioned above, the $k$-colorable $G_{n,p}$ class appears maximally hard for all known algorithms with respect to a phase transition defined by $n, p$ and $k$, where $k \approx n/\log_b n$ and $b = 1/(1 - p)$. The Hamiltonian cycle $G_{n,m}$ class on the other hand does not appear maximally hard for any value of $m$. In fact, for large $n$ our algorithm almost never takes more than $O(n)$ backtrack nodes and $O(nm)$ running time.

We will use a much weaker requirement and say an instance is *quadratically hard* if it requires at least $n^2$ search nodes by the backtrack algorithm described in section 3. Note that $\Omega(n^2)$ search nodes would take our algorithm $\Omega(n^3)$ time. For practical reasons, we will also use a weaker definition for robustness, and say that an instance is *robustly quadratically hard* if our algorithm uses at least $n^2$ search nodes when the iterated restart feature is used with a multiplying factor of 2. (See section 3 for program details). We say a class is *minimally hard* if there is some constant $\epsilon > 0$ such that the probability of a hard instance is at least $\epsilon$ as $n \to \infty$.

In Section 4 we examine $G_{n,m}$ random graphs using our backtrack algorithm on graphs of up to 1500 vertices. The empirical evidence we collect suggests that in contrast to the graph coloring situation, the Hamiltonian cycle $G_{n,m}$ class is not minimally quadratically hard, even for $m$ at or near the phase transition, and even if we drop our minimal robustness requirement.

Note that we do not dispute the claim that hard instances are more likely at the phase transition than at other values of $m$, but rather claim that even at the transition the probability of generating a hard instance rapidly goes to zero with increasing $n$.

## 2.2 Random Graph Theory and the Phase Transition

These results are not unexpected when one reviews the theoretical work on this graph class. Since asymptotically the graph becomes Hamiltonian when an edge is added to the last degree 1 vertex (Bollobás, 1984), any algorithm that checks for a minimum degree $\leq 2$ will detect almost all non-Hamiltonian graphs. When the graph is Hamiltonian, various researchers (Angluin & Valiant, 1979; Bollobás, Fenner, & Frieze, 1987) have proven the existence of randomized heuristic algorithms which can almost always find a Hamiltonian cycle in low-order polynomial time. In particular, it is shown (Bollobás et al., 1987) that there is a polynomial time algorithm HAM such that

$$\lim_{n \to \infty} \Pr\left(\text{HAM finds a Hamilton cycle}\right) = \begin{cases} 0 & \text{if } c_n \to -\infty \\ e^{-e^{-2c}} & \text{if } c_n \to c \\ 1 & \text{if } c_n \to \infty \end{cases}$$

where $m = n/2(\ln n + \ln \ln n + c_n)$.

Furthermore, as the authors point out, this is the best possible result in the sense that this is also the asymptotic probability that a $G_{n,m}$ graph is Hamiltonian, and is the probability that it has a minimum degree of 2. In other words, the probability of finding a cycle is the same as the probability of one existing. Given that it is trivial to check the minimum vertex degree of a graph, this does not leave much room for the existence of hard instances (for HAM and similar algorithms).

Another relevant theoretical result is that there is a polynomial time algorithm which with probability going to one, finds some Hamiltonian cycle when a graph has a hidden

Hamiltonian cycle together with extra randomly added edges(Broder, Frieze, & Shamir, 1994). For the algorithm to work, the average degree of a vertex needs only be a constant. They claim the result can be easily extended to the case that the average degree is a growing function of $n$. This is another indication that Hamiltonian graphs near the phase transition will be easy to solve by some algorithm.

For a non-Hamiltonian graph to be hard for an algorithm it must contain a feature preventing the formation of a Hamiltonian cycle which the algorithm cannot easily detect. Suppose a backtrack algorithm does not check for vertices of degree one. The algorithm may then require exponential backtrack before determining the non-Hamiltonicity of the graph, since the only way it can detect this is by trying all possible paths and failing. However, degree one vertices are easily detectable, and so are not good indicators of hard instances. They also disappear at the phase transition.

Similarly, an algorithm might not check for articulation points, and as a result waste exponential time on what should be easy instances. As $n \to \infty$, the probability of an articulation point existing (in $G_{n,m}$) goes to zero as fast as the probability of the existence of a vertex of degree less than two. Other features can lead to non-Hamiltonicity of course, such as $k$-cuts that leave $k+1$ or more components (Bondy & Murty, 1976), and these could require time proportional to $n^k$ to detect. Under the assumption that NP$\neq$CO-NP there must also exist a set of non-Hamiltonian instances which have no polynomial proof of their status.

However, it seems that at the phase transition the larger the feature the less likely it is to occur. In fact, the theoretical results summarized above indicate this must happen. Although we know hard graphs exist, and we may expect these localized types of hard graphs to be more frequent near the phase transition than elsewhere when using $G_{n,m}$ to generate instances, we also expect the probability of such instances to go to zero as $n$ increases.

## 3. An Overview of our Backtrack Algorithm

Our backtrack algorithm comes from Vandegriend (1998), and is based upon prior work on backtrack Hamiltonian cycle algorithms (Kocay, 1992; Martello, 1983; Shufelt & Berliner, 1994). It has three significant features which we will discuss. First, it employs a variety of pruning techniques during the search that delete edges that cannot be in any Hamiltonian cycle. This pruning is usually based upon local degree information. Second, before the start of the search the algorithm performs initial pruning and identifies easily detectable non-Hamiltonian graphs. The third feature is the use of an iterated restart technique. Additionally, the program provides the opportunity to order the selection of the next vertex during path extension using either a low degree first ordering, a high degree first ordering, or a random ordering. We normally use the low degree first ordering.

At each level of the search, after adding a new vertex to the current path, search pruning is used. The pruning identifies edges that cannot be in any Hamiltonian cycle and removes them from the graph. (Note that if the algorithm backtracks, it adds the edges deleted at the current level of the search back to the graph.) The first graph configuration that the pruning looks for is a vertex $x$ with 2 neighbours $a, b$ of degree 2. Since the edges incident on $a$ and $b$ must be used in any Hamiltonian cycle, the other edges incident on $x$ can be deleted. The second graph configuration that the pruning looks for is a path $P = (v_1, \ldots, v_k)$ of forced edges (so $v_2 \ldots v_{k-1}$ are of degree 2). If $k < n$ then the edge $v_1, v_k$ cannot be in any Hamiltonian cycle and can be deleted. If as a result of pruning, the degree of any vertex drops below 2, then no Hamiltonian cycle is possible and the algorithm must backtrack. The use of these operators may yield new vertices of degree 2 and therefore the pruning is iterated until no further changes occur.

A pruning iteration takes $O(n)$ time to scan the vertices to check for vertices with two degree 2 neighbors, and $O(n)$ time to extend all forced degree two paths. Since the iterations terminate unless a new vertex of degree two is created, at most $n$ iterations can occur. At most $O(m)$ edges can be deleted. On backing up from a descendant, the edges are replaced ($O(m)$) and the next branch is taken. Thus, an easy upper bound on the pruning time for a node searching from a vertex of degree $d$ is $O(d(n^2 + m))$, but this is overly pessimistic. Note that along any branch from the root of the search tree to a leaf, at most $n$ vertices can be converted to degree 2. Also note that along each branch each edge can be deleted at most once. If the degree is high we seldom take more than a few branches before success. The implementation is such that when several vertices have two neighbors of degree two at the beginning of an iteration, all redundant edges are removed in a single pass taking time proportional to $n$ plus the number of edges removed and checked. In practice, on $G_{n,m}$ graphs it typically takes $O(n+m)$ time per search node on very easy Hamiltonian instances as evidenced by CPU measurements, with harder instances taking at most twice as long per search node.

Before the start of the recursive search, our algorithm prunes the graph as described above. Then the algorithm checks to see if the graph has minimum degree $\geq 2$, is connected, and has no cut-points. If any of these conditions are not true, then the graph is non-Hamiltonian and the algorithm is finished.

Some non-Hamiltonian instances may be very easy or very hard to detect, depending on which vertex the algorithm chooses as a starting point. In these cases local features exist that could be detected if the algorithm starts near them, but otherwise the algorithm may backtrack many times into the same feature without recognizing that only the feature matters. The seemingly hard instance on $G_{n*}$ for $n = 100$ discussed in Section 4.2 is such a case. This is one type of "thrashing," and is a common problem in backtracking algorithms. For example, Hogg and Williams (1994) noticed a sparse set of very hard 3-coloring problems that were not at the phase transition. Baker (1995) showed that these instances were most often hard as a result of thrashing, and that they could be made easy by backjumping or dependency-directed backtracking.

To improve our algorithm's average performance we use an iterated restart technique. The idea is to have a maximum limit $M$ on the number of nodes searched. When the maximum is reached, the search is terminated and a new one started with the maximum increased by a multiple $k$ (so $M_{i+1} = kM_i$). Initially, $M = kn$. In our experiments, we used $k = 2$. By incrementing the search interval in this way, the algorithm will eventually obtain a search size large enough to do an exhaustive search and thus guarantee eventual completion. The total search will never be more than double the largest size allocated.

Although random restarts are sometimes effective on non-Hamiltonian graphs, they are more frequently effective on Hamiltonian instances. During search, as edges are added to the set of Hamiltonian edges, the net effect is to prune edges from the graph. For a Hamiltonian graph to be hard, the algorithm must select some set of edges which causes the reduced graph to become non-Hamiltonian, and this non-Hamiltonian subgraph must itself be hard to solve. With iterated restart, for the instance to remain hard the algorithm must make such mistakes with high probability. As a result, we expect fewer hard Hamiltonian instances.

Random restarts are an integral part of randomized algorithms (Motwani & Raghavan, 1995) and are used frequently in local search and other techniques to escape from local optima (Johnson, Aragon, McGeoch, & Schevon, 1991; Langley, 1992; Selman, Levesque, & Mitchell, 1992; Gomes, Selman, & Kautz, 1998). Further discussion of the impact of restarts can be found in the analysis of the experiments on $G_{n,m}$ graphs in Section 4.

The algorithm also provides for the possibility of checking for components and cut vertices during recursive search after the pruning is completed at each search node. The overhead of this extra work is $O(n+m)$ per search node and rarely seems to pay off. Except where noted these checks were not used in this study.

The experimental results reported in the remaining sections were run on a variety of machines, the fastest of which is a 300 MHZ Pentium II. All CPU times reported are either from this machine, or adjusted to it using observed speed ratios on similar tests. Our algorithm terminated execution after 30 minutes[2]. Experimental results are frequently reported as the ratio of the number of search nodes over the number of vertices. This node ratio is used because we feel it provides a better basis for comparing results across different graph sizes, since many of our results are $O(n)$. Note that the number of search nodes is calculated as the number of recursive calls performed.

We used several different methods of verifying the correctness of our algorithm and our experimental results. The algorithm was independently implemented twice, and performs automatic verification of all Hamiltonian cycles found. We performed multiple sets of experiments on generalized knight's circuit graphs and compared the results (graph Hamiltonian or not) to our theoretical predictions. Initial sets of experiments on $G_{n,m}$ graphs and Degreebound graphs were executed using two different pseudo-random number generators, and were repeated multiple times. Our source code is available as an appendix.

## 4. $G_{n,m}$ Random Graphs

We consider random graphs of 16 to 1500 vertices with $m = \overline{d}n/2$. From previous work (Cheeseman et al., 1991; Komlós & Szemerédi, 1983) we expect the phase transition to occur when $\overline{d} \approx \ln n + \ln \ln n$. Thus we specify the constraint parameter (or degree parameter) $k = \overline{d} / (\ln n + \ln \ln n)$.

### 4.1 $G_{n,m}$ Using Restart

For the premiere experiment, we generate $G_{n,m}$ graphs with number of vertices $n = 16 \ldots 96$ in steps of 4, $n = 100 \ldots 500$ in steps of 100, $n = 1000$ and $n = 1500$. For each size $n$, the degree parameter $k$ ranges from $0.5 \ldots 2.0$ (step size 0.01 from $k = 1.00 \ldots 1.20$, step size 0.10 for other ranges of $k$). We generate 5000 graphs for each data point and execute our backtrack algorithm once on each graph. This is a grand total of 4.76 million graphs, of which 1.19 million are of 100 or more vertices.

---

2. Since the time limit of 30 minutes is at least two orders of magnitude greater than the typical running time, the limit is rarely used. On slower machines this limit was increased. The Knight's tour graphs reported in Section 6 were run on a slower machine with a 30 minute time limit, although some instances were run much longer.

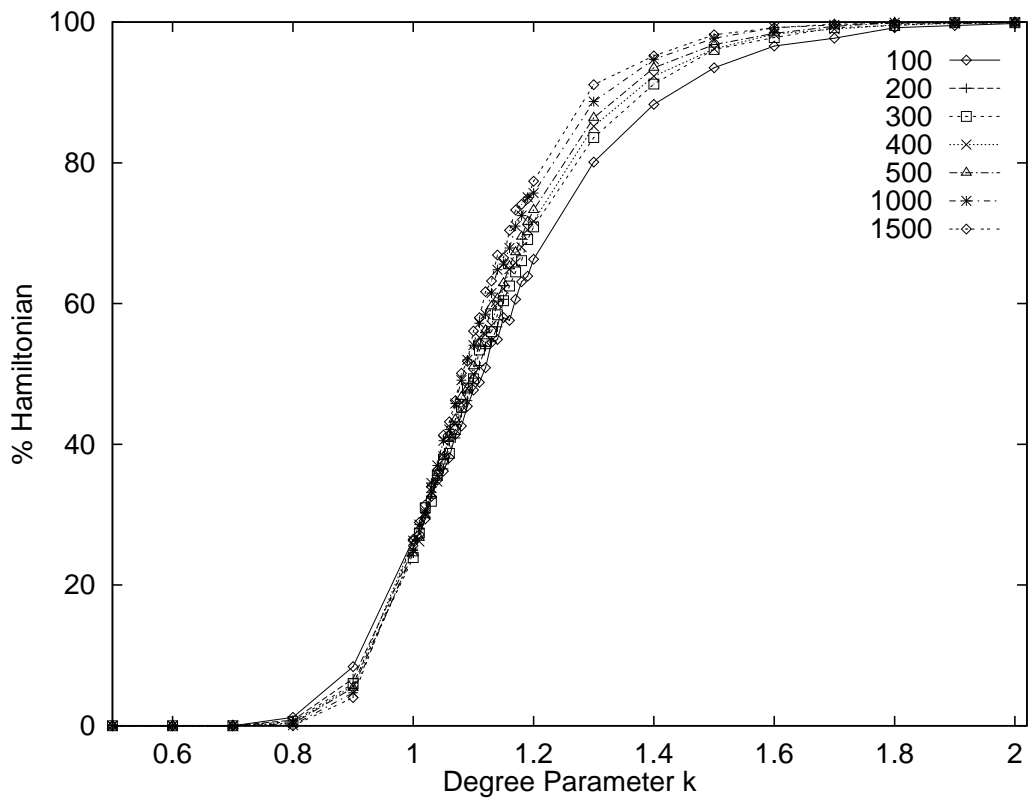

Figure 1: % of Hamiltonian graphs as a function of graph size and degree parameter for $G_{n,m}$ graphs.

We use the pruning described in section 3, check for components and articulation points after the initial pruning, and use iterated restart with a multiplicative factor of 2. We do not check for components or articulation points during the recursive search.

We expect the phase transition for biconnectivity to be very similar to the phase transition for Hamiltonicity (Cheeseman et al., 1991) and we expect the phase transition for minimum degree greater than 1 to be almost identical to the phase transition for Hamiltonicity (Bollobás, 1984; Komlós & Szemerédi, 1983). Our experimental results matched these expectations very closely. For the larger graphs of 100 to 1500 vertices, the percentage of Hamiltonian graphs is plotted against the degree parameter in Figure 1. We found that the 50% point at which half the graphs are Hamiltonian occurs when the degree parameter $k \approx 1.08 - 1.10$. More interestingly, all the curves pass close to a fixed point near $k = 1$, and it seems they are approaching a vertical line at this point. That is, they appear to be converging on $k \approx 1$ as a phase transition, precisely as theory predicts.

| $n$     | 100   | 200   | 300   | 400   | 500   | 1000  | 1500  |
|---------|-------|-------|-------|-------|-------|-------|-------|
| Nodes   | $7.5n$ | $7.0n$ | $3.3n$ | $7.0n$ | $3.4n$ | $3.3n$ | $7.0n$ |

Table 1: Maximum Search Nodes on $G_{n,m}$ for Large $n$

All graphs were solved, that is were either determined to be non-Hamiltonian, or a Hamiltonian cycle was found. We are primarily interested in asymptotic behavior, since theories concerning the relation of the phase transition to hard regions are necessarily asymptotic in nature. For graphs of 100 vertices or more, the longest running time was under 11 seconds, on a graph of 1500 vertices using 10,500 (or $7.0n$) search nodes to find a Hamiltonian cycle.

All of the 549,873 non-Hamiltonian graphs in this range were detected during the initial pruning of the graph, and thus no search nodes were expanded. Of the 640,127 Hamiltonian $G_{n,m}$ graphs, the vast majority ( 629,806 or 98.3%) used only $n$ search nodes, which means that the algorithm did not need to backtrack at all[3]. No quadratically hard graphs were found in this range. Table 4.1 lists the maximum number of search nodes expressed as a factor of $n$ to illustrate the linearity of the search tree.

These results appear to differ from those of Frank et al. (1998), who found graphs which took orders of magnitude more search nodes to solve. (Their hardest graph took over 1 million nodes.) We believe this is due to two factors. Firstly, the algorithm used to generate the results in their paper did not do an initial check for biconnectivity nor did it use all of the pruning techniques used in our algorithm. Secondly and more importantly, on the small random graphs they used ($\leq 30$ vertices) the probability of obtaining certain hard configurations (such as biconnected and non-Hamiltonian or non-biconnected and minimum degree $\geq 2$) is much higher than when $n$ is larger, as we discussed in section 2.2.

The experiments on small $G_{n,m}$ graphs (between 16 and 96 vertices) confirm this conjecture. In this case we do find a small number of quadratically hard graphs, and a few very hard graphs. We consider for purposes of this paper, that a very hard graph on less than 100 vertices is any that takes at least 100,000 search nodes to solve. The very hard graphs from this set of runs are given in Table 4.1.

Note that the very hardest took less than two minutes to solve, making our designation of "very hard" questionable. Also, note that the smallest graph in this set has 36 vertices, somewhat larger than the 30 vertex examples found by Frank et al. (1998). This is likely because we do articulation point checking initially and better pruning. Finally, all of these very hard graphs are non-Hamiltonian, and all occur in classes that produce less than 50% Hamiltonian graphs. The hardest Hamiltonian graph in contrast required only 19,318 search nodes, on a graph of 68 vertices with degree parameter 0.9.

In Figure 2 we plot the number of graphs that are quadratically hard for these small graphs. For $n$ from 68 to 92, all non-Hamiltonian graphs were detected during initial pruning. One non-Hamiltonian graph at $n = 96$ required search ($254.1n$ nodes). Notice that the number of quadratically hard Hamiltonian graphs is far less than the number of quadratically hard non-Hamiltonian graphs, and peaks for larger $n$. This is in accordance with the discussion of random restarts in Section 3.

---

3. With 5% error in this measurement, this means that the algorithm might have backtracked over a maximum of $0.05n$ search nodes.

| Vertices | Degree Parameter | Seconds | Search Nodes | Ratio |
|---|---|---|---|---|
| 36 | 1.11 | 94.7 | 1179579 | 32766.1 |
| 40 | 1.00 | 36.5 | 638946 | 15973.6 |
| 40 | 1.07 | 18.7 | 327603 | 8190.1 |
| 44 | 1.00 | 12.3 | 156694 | 3561.2 |
| 44 | 1.04 | 20.0 | 293664 | 6674.2 |
| 48 | 1.02 | 91.2 | 1280135 | 26669.5 |
| 48 | 1.09 | 107.0 | 1243647 | 25909.3 |

Table 2: The Hardest Small Graphs

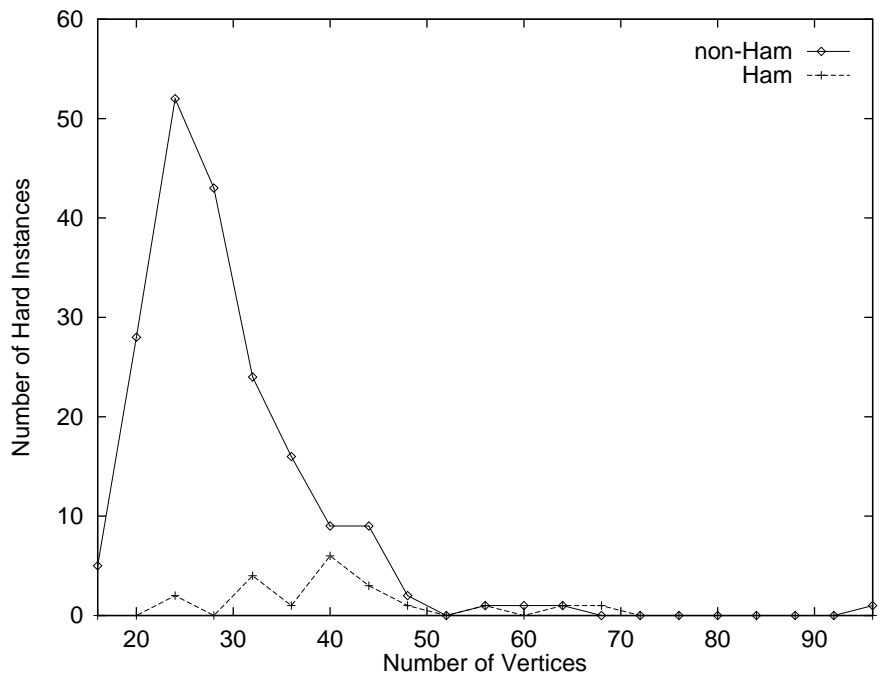

Figure 2: The Number of Quadratically Hard Graphs for Small $n$.

We ran additional tests for $n$ from 32 to 54 in steps of 2, with the degree parameter ranging from 0.96 to 1.16 with step size 0.01, generating 5000 graphs at each point. In this case, we invoked articulation point checking at each search node. Again all graphs were solved without timing out, and some very hard graphs were found, all of them non-Hamiltonian. One 50 vertex graph required 9,844,402 search nodes, and required close to 20 minutes to solve. It is unclear whether the extra checking helped; the smallest graph requiring at least 100,000 nodes had 32 vertices, while the smallest requiring over a million had 40 vertices. Overall, the results were very similar to the first set of experiments on small graphs.

### 4.2 $G_{n*}$ Using Restart

Clearly, the more edges we add to a graph, the more likely it is to be Hamiltonian. It also seems that once a graph is Hamiltonian, adding more edges makes it less likely to be hard. In an attempt to find hard graphs for larger $n$, we modified the $G_{n,m}$ generator so that instead of adding a fixed number of edges, it instead added edges until every vertex has degree at least two, and then stops. In a sense this produces graphs exactly on the $G_{n,m}$ phase transition, since a minimum degree of two is the condition that asymptotically distinguishes Hamiltonian from non-Hamiltonian graphs with high probability. We refer to this distribution as the $G_{n*}$ model.

Initially we ran 1000 graphs with this generator for $n$ from 100 to 500, but no hard instances were found. We increased the search to 10,000 graphs at each $n$, and included a search at $n = 1000$. Out of all these graphs, we found one very hard graph on 100 vertices. Even after a second attempt using more than 26 million search nodes, it was still unsolved. Doing post-mortem analysis, we checked for cut sets of size 2 and 3 that would leave 3 or 4 (or more) components and found none. We also checked the pruned graph using the odd degree test mentioned in Section 5.3, but this too failed to show it is non-Hamiltonian. Finally, we set up our fast machine with unlimited time and no restarts. Three search nodes and less than 0.1 seconds later it was proven non-Hamiltonian.

Detailed analysis (see the appendix) shows that the graph has a small feature that is easily detected when one of a few starting points is selected. Because we use an exponentially growing sequence of searches, we only use a few restarts. In a test of 100 random starts with a 3 second time limit 7 trials succeeded, using from 2 to 5 search nodes each to prove the graph non-Hamiltonian.

We also ran 10,000 $G_{n*}$ graphs at each even value of $n$ from 16 to 98. The smallest instances requiring at least 100,000 search nodes were at $n = 50$. Only 5 graphs requiring more than a million nodes were found for $n < 100$, two at $n = 62$, one at $n = 70$ and two at $n = 98$. Two of these (one at 62, one at 98) initially timed out, but were solved in second attempts in about 1/2 hour. Neither was susceptible to an attack by 100 restarts as on the 100 vertex graph.

Table 4.2 shows the number of non-Hamiltonian graphs for each $n \geq 100$. All of these except the one mentioned above were detected during initial pruning. The remaining graphs were all easily shown to be Hamiltonian, with a maximum search ratio of 7.0.

Clearly the probability of non-Hamiltonian graphs drawn from $G_{n*}$ is decreasing with $n$. It seems likely that the probability of hard instances is also going to zero.

| $n$ | 100 | 200 | 300 | 400 | 500 | 1000 |
|---|---|---|---|---|---|---|
| Non-Ham | 154 | 56 | 29 | 20 | 15 | 3 |

Table 3: Number of Non-Hamiltonian Graphs from $G_{n*}$

| $n$ | $k = 1.00$ | $k = 1.50$ | $k = 2.00$ |
|---|---|---|---|
| 500 | 0.20 | 0.20 | 0.21 |
| 1000 | 0.43 | 0.50 | 0.60 |
| 1500 | 0.68 | 0.80 | 0.87 |

Table 4: CPU Seconds per 1000 Search Nodes for $G_{n,m}$ Graphs

### 4.3 $G_{n,m}$ Without Using Restart

We wanted to know how important the restart feature is asymptotically. We ran 1000 $G_{n,m}$ graphs for $n$ from 100 to 1500, for each of the parameter settings in the premiere experiment, but this time using the backtrack algorithm *without* the iterated restart feature. As before, all non-Hamiltonian instances were detected during initial pruning. One quadratically hard Hamiltonian graph was found at $n = 300$, with degree parameter 1.20, which required 163,888, or $1.82n^2$ search nodes and took 28.5 seconds. A few other graphs were nearly quadratic, for example on $n = 1500$ there were 4 graphs that required $0.15n^2$, $0.19n^2$, $0.36n^2$ and $0.47n^2$ search nodes. It seems that asymptotically, even in the absence of iterated restarts, the $G_{n,m}$ class does not provide hard instances with high probability.

### 4.4 $G_{n,m}$ Summary

Based on a set of timing runs, we present in Table 4.4 an indication of how running time per search node increases with the number of vertices $n$ and degree parameter $k$. Because the times are usually so short, we cannot get reliable numbers for $n < 500$. The times shown are for the evaluation of 1000 search nodes, and are averaged (total CPU divided by total nodes searched) over graphs that were solved in less than $1.1n$ search nodes. For instances that require significantly more search nodes, the time per 1000 nodes seems to increase somewhat, but there are so few examples for large $n$ that we are unable to provide exact estimates. For $n = 1500$[4], the average time per 1000 nodes for instances requiring more than $2n$ search nodes is 0.89 seconds at $k = 1.00$, 1.04 at $k = 1.50$ and 1.31 at $k = 2.00$. Note that this includes at least one instance that took $7n$ search nodes. This table indicates that the growth is approximately linear in $n + m$.

The experimental evidence clearly indicates that $G_{n,m}$ random graphs are asymptotically extremely easy everywhere, despite the existence of a phase transition. Our results temper the findings of the various researchers (Cheeseman et al., 1991; Frank et al., 1998; Frank & Martel, 1995) studying phase transitions and the Hamiltonian cycle problem. Cheeseman et al.'s explanation of their observed increase in difficulty near the phase transition was that "on the border [between the regions of low and high connectivity] there are many almost Hamiltonian cycles that are quite different from each other ... and these numerous local minima make it hard to find a Hamiltonian cycle (if there is one). Any search procedure based on local information will have the same difficulty." (Cheeseman et al., 1991). Unfortunately, while their observations were accurate, their observed hardness was due to their algorithms and the limited size of the graphs tested, not to intrinsic properties of the Hamiltonian cycle problem with respect to the phase transition on $G_{n,m}$ graphs. We have shown that an efficient backtrack algorithm finds the phase transition region of $G_{n,m}$ graphs easy in general.

---

4. $n = 1500$ is the only value of $n$ for which we have at least one instance requiring $\geq 2n$ search nodes at each of the three values of $k$. The times for 1000 and 1500 come from separate runs on 1000 graphs per sample point.

## 5. Degreebound Graphs

Intuitively, the reason that it is so hard to generate a hard instance from $G_{n,m}$ is that by the time we add enough edges to make the minimum degree two, the rest of the graph is so dense that finding a Hamiltonian cycle is easy. Alternatively, we see that to create a non-Hamiltonian property or feature, we must have regions of low degree, while at the same time meeting the minimal requirements that make the instance hard to solve. This problem can be characterized as one of high variance of vertex degrees. The only region where we get even a few hard graphs from $G_{n,m}$ is when $n$ is small enough that the average degree is also low.

To avoid the consequences of this degree variation, in this section we use a different random graph model $G_n(d_2 = p_2, d_3 = p_3, \ldots)$ for which $n$ is the number of vertices and $d_i = p_i$ is the percentage of vertices of degree $i$. As an example $G_{100}(d_2 = 50\%, d_3 = 50\%)$ represents the set of graphs of 100 vertices in which 50 are of degree 2 and 50 are of degree 3. We refer to a graph generated under this model as a Degreebound graph. In this paper we only consider graphs whose vertices are of degree 2 or 3.

It is quite difficult to generate all graphs with a given degree sequence with equal probability (Wormald, 1984). Instead, we adopt two variations which generate graphs by selecting available edges. In each case each vertex is assigned a free valence equal to the desired final degree. In version 1 pairs of vertices are selected in random order, and added as edges if the two vertices have at least one free valence each. This continues until either all free valences are filled (a successful generation) or all vertex pairs are exhausted (a failure). If failure occurs, the process is repeated from scratch. Initial tests indicate about 1/3 of the attempts fail in general. For efficiency reasons, in the implementation an array of vertices holds each vertex once. Pairs of vertices, $v, w$ are selected at random from the array and if $v \neq w$, and $(v, w)$ is not already an edge, then $(v, w)$ is added as an edge, and the free valence of each of $v$ and $w$ is reduced by one. When the free valence of a vertex is zero, the vertex is deleted from the array. This step is repeated until only a small number (twice the maximum degree) of vertices remains, and then all possible pairs of the remaining vertices are generated and tested in random order.

In version 2 an array initially holds each vertex $v$ $deg[v]$ times. Pairs of vertices are randomly selected, and if not equal and the edge does not exist, then the edge is added, and the copies of the two vertices are deleted from the array. This is repeated until the array is empty, or 100 successive attempts have failed to add an edge. The latter case is taken as failure, and the process is repeated from scratch. This method seldom fails.

Neither of these two methods guarantees a uniform distribution over the graphs of the given degree sequence. For example, given the degree sequence on five vertices $\{1, 1, 2, 2, 2\}$, there are seven possible (labeled) graphs. One consists of two components, an edge and a triangle. The other six are all four paths; thus all six are isomorphic to one another. Of the 10! permutations of the pairs of vertices, 564,480 generate the graph on two components, while for each four path there are 322,560 distinct permutations. The remaining permutations (31.2 %) do not yield a legal graph. Thus, the first graph is 1.75 times as likely as any of the other six. Of course, a four path (counting all isomorphic graphs) is 3.428 times as likely as the two-component graph.

On the other hand, a version 2 test program (not our generator which prohibits degree one vertices) consistently generated the first graph about 8%–10% more often than any of the others, based on several million random trials.

## 5.1 Experimental Results on Degreebound Graphs

We test graphs of $100 \ldots 500$ vertices (step size 100) 1000 and 1500 vertices with the mean degree varying from $2.6 \ldots 3.0$ (step size of 0.01 from 2.75 to 2.95, step size of 0.05 elsewhere). We generate 1000 graphs for each data point, execute our algorithm once on each graph, and collect the results. This test was repeated for each of the two versions.

Figure 3 shows the percentage of graphs which are Hamiltonian as the mean degree and graph size varies[5]. There is a clear transition from a mean degree of 2.6 (near 0% chance of a Hamiltonian cycle) to a mean degree of 3 (for which Robinson and Wormald, 1994 predict an almost 100% chance of a Hamiltonian cycle on uniformly distributed graphs). For a phase transition, we would expect the slope to grow steeper as the graph size increases. Figure 3 shows this increase in steepness.

Note that the double points on the curve for $n = 100$ are due to unavoidable discretization. Since the total degree of a graph must be even, when the generators detect that the total degree specified is odd, one of the minimum degree vertices is selected and its degree incremented. Thus, for example, whether the fraction of degree 3 vertices specified is 0.81 or 0.82, the number of degree three vertices is 82. Discretization effects also occur for $n = 300$, 500 and 1500, but with lessened impact.

In Table 5.1 we summarize the observed hard instances from these graphs. We note that several instances exceeded our time bounds, and although these are certainly at least quadratically hard, they are not included in the quadratically hard instances. The frequency of hard instances appears to be decreasing with $n$ on these graphs. In particular there are no quadratically hard non-Hamiltonian instances over 1000 vertices, except those that are too hard to solve with our program.

Interestingly, there turns out to be an $O(n + m)$ time test which shows that most of the unresolved instances are non-Hamiltonian. This test is described briefly in Section 5.3. We implemented the test as a separate program and tested each of the unresolved graphs, with the results indicated in the last column of Table 5.1. The remaining five graphs remain unresolved. If this test were included in the initial pruning of our program, then the instances enumerated in the last column of Table 5.1 would all be solved (proven non-Hamiltonian) without search.

---

5. For these graphs, the mean degree is 2.0 plus the fraction of degree 3 vertices.

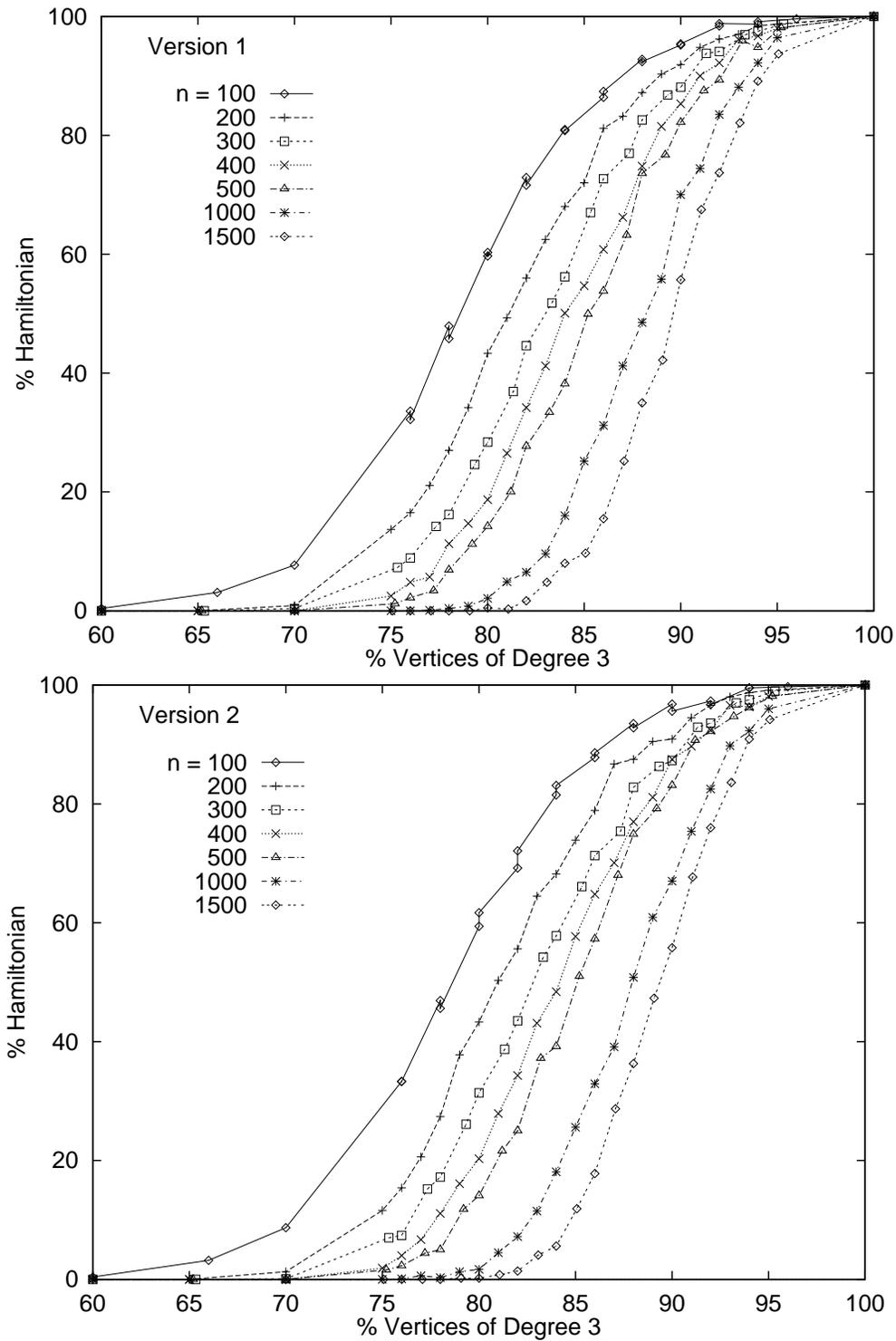

Figure 3: % of Hamiltonian graphs for Degreebound Graphs.

| Version 1 | | | | |
|---|---|---|---|---|
| Number of | Quadratically Hard | | Timed Out | |
| Vertices | No HC | HC | Total | No HC |
| 100 | 5 | 0 | 0 | 0 |
| 200 | 18 | 0 | 3 | 3 |
| 300 | 8 | 0 | 11 | 10 |
| 400 | 1 | 0 | 14 | 14 |
| 500 | 0 | 0 | 14 | 14 |
| 1000 | 0 | 0 | 7 | 7 |
| 1500 | 0 | 1 | 6 | 6 |
| Version 2 | | | | |
| Number of | Quadratically Hard | | Timed Out | |
| Vertices | No HC | HC | Total | No HC |
| 100 | 5 | 0 | 0 | 0 |
| 200 | 9 | 0 | 6 | 5 |
| 300 | 10 | 0 | 13 | 13 |
| 400 | 3 | 0 | 11 | 11 |
| 500 | 1 | 1 | 10 | 9 |
| 1000 | 0 | 1 | 6 | 4 |
| 1500 | 0 | 0 | 6 | 6 |

Table 5: Number of Hard Graphs for Degreebound Graphs

Thus, although these classes may provide a small rate of hard instances for our current program, it is not clear they are even minimally hard. Furthermore, it appears there exist simple improvements to our program that would eliminate most of these hard instances.

In Figure 4 we illustrate the distribution of the graphs that timed out. The other quadratically hard graphs had similar distributions. About all that can be concluded is that the hard instances seem to be distributed over a mean degree range from 2.78 to 2.94.

The backtrack program is a little faster on Degreebound graphs than on $G_{n,m}$ graphs, as we would expect given fewer total edges. For 1500 vertices, the times per 1000 search nodes ranged from 0.27 seconds for the easiest (no backtrack) instances to 0.56 seconds for the harder ones.

## 5.2 Analysis of Degreebound Graphs

An analysis of the Degreebound graph class led us to conjecture that the prime factor determining the Hamiltonicity of a graph was whether or not the graph had a degree 3 vertex with 3 neighbours of degree 2. We label this a 3D2 configuration (or a 3D2 event). A graph with a 3D2 configuration is non-Hamiltonian. The following informal analysis provides evidence for our conjecture.

Let $E(n, \epsilon)$ represent the expected number of 3D2 configurations in a graph with $n$ vertices. Let $D_2 = \epsilon n$ be the number of degree 2 vertices and $D_3 = (1 - \epsilon)n$ the number of degree 3 vertices. Note that the mean degree $\overline{d} = \frac{2D_2 + 3D_3}{n} = \frac{2\epsilon n + 3n(1 - \epsilon)}{n} = 3 - \epsilon$. Assuming equal probability of all combinations,

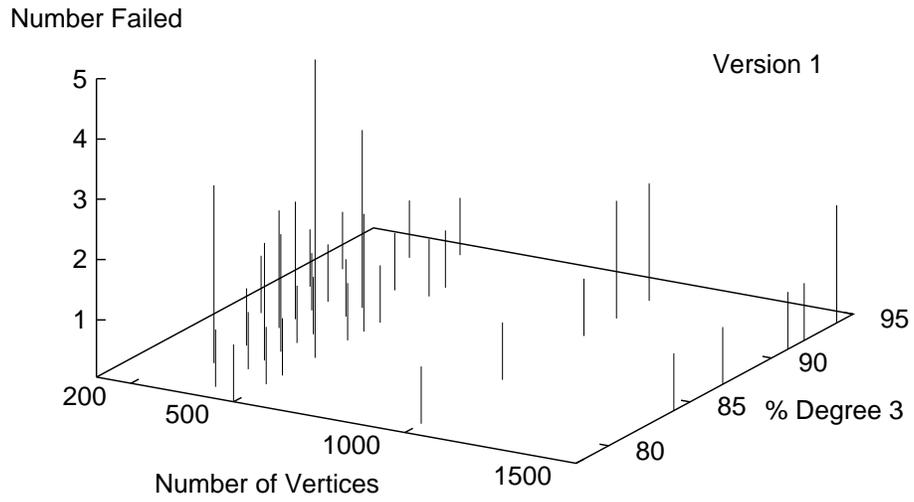

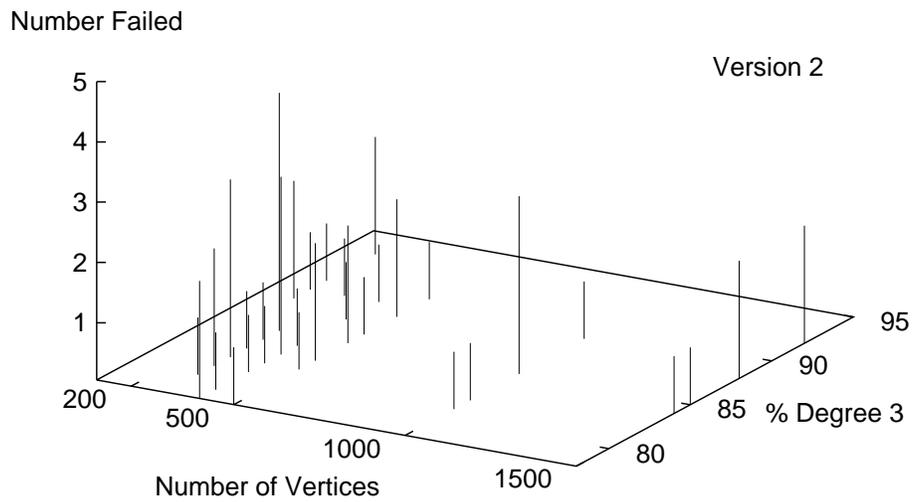

Figure 4: Distribution of Timed Out Instances for Degreebound graphs

| # of | Mean Degree for 50% HC Point | |
|---|---|---|
| Vertices | Experimental | Theoretical |
| 100 | 2.78 | 2.78 |
| 200 | 2.81 | 2.82 |
| 300 | 2.83 | 2.85 |
| 400 | 2.84 | 2.86 |
| 500 | 2.85 | 2.87 |
| 1000 | 2.88 | 2.90 |
| 1500 | 2.90 | 2.91 |

Table 6: Experimental and approximate theoretical values for the location of the 50% Hamiltonian point for Degreebound graphs of various sizes.

$$E(n, \epsilon) = \frac{D_3 \binom{D_2}{3}}{\binom{n-1}{3}} = \frac{n(1-\epsilon)\binom{\epsilon n}{3}}{\binom{n-1}{3}} = \frac{n(1-\epsilon)(\epsilon n)(\epsilon n - 1)(\epsilon n - 2)}{(n-1)(n-2)(n-3)}$$

We restrict ourselves to the asymptotic case $(n \to \infty)$ which gives us

$$E(n, \epsilon) \approx \frac{n(1-\epsilon)(\epsilon n)^3}{n^3} \approx n(1-\epsilon)\epsilon^3$$

When $E(n, \epsilon) \to 0$, the probability of having configuration 3D2 approaches 0. We want to find $\epsilon$ for which $n(1-\epsilon)\epsilon^3 \to 0$ as $n \to \infty$. This occurs when $\epsilon = o(n^{-1/3})$. Since a Hamiltonian cycle cannot exist if $E(3D2) > 0$, this tells us that the phase transition asymptotically occurs when the mean degree equals 3. Asymptotically, Degreebound graphs with $\overline{d} < 3$ are expected to be non-Hamiltonian while Degreebound graphs with $\overline{d} > 3$ are expected to be Hamiltonian (ignoring other conditions). This agrees with results of Robinson and Wormald (1994) who proved that almost all 3-regular graphs are Hamiltonian.

If we let $\epsilon = n^{-1/3}$ this gives us $E(n, \epsilon) \approx 1$. Substituting this equation in our expression for mean degree gives us $\overline{d} = 3 - n^{-1/3}$. Table 5.2 lists mean degrees for different values of $n$ using this formula along with our experimentally determined values for the point where 50% of the graphs are Hamiltonian. They are remarkably similar. This suggests that the 3D2 configuration is the major determinator of whether a Degreebound graph will be Hamiltonian or not. Minor effects (which we have ignored) come from propagation of deleted edges while pruning and other less probable cases such as those mentioned in Section 5.3. Since the 3D2 configuration is detected by our algorithm before the search is started, this also implies that the phase transition will be easy for our algorithm, since most non-Hamiltonian graphs are instantly detected. This matches our experimental observations.

## 5.3 A Non-Hamiltonicity Test for Sparse Graphs

While preparing the final version of this paper, we observed that in the 3D2 configuration we could replace the vertex of degree three with a component of several vertices. In general, if there are three vertices of degree two that form a minimal cut then the graph is non-Hamiltonian. In fact, we can replace the three vertices by a minimal cut of any odd number $c$ of degree 2 vertices, and the claim of non-Hamiltonicity remains true.

Checking all possible subsets of size $c$ would be very expensive, but fortunately there is an even more general condition that includes all of these as special cases and can be tested in linear (i.e $O(n+m)$) time. Let $F$ be a set of edges that are forced to be in any Hamiltonian cycle if one exists. For example, edges incident on a vertex of degree two are forced. Let $G' = G - F$ be the graph formed by deleting the forced edges from $G$. Let $C_1 \dots C_h$ be components of $G'$, and define the *forced degree* of component $C_i$ to be the number of end points of forced edges (from $F$) in $C_i$. If any component has an odd forced degree, then $G$ is non-Hamiltonian.

The proof of correctness of this test is simple. Observe that if there is a Hamiltonian cycle in $G$ then while traversing the cycle each time we enter a component, there must be a corresponding exit. Since the forced edges act as a cut set (that separates the components), they are the only edges available to act as entries and exits to a component. All forced edges must be used. Therefore, if there is a Hamiltonian cycle there must be an even number of forced edges connecting any component to other components, each contributing one to the forced degree of the component. Each forced edge internal to (with both end points in) a component contributes two to the forced degree, so if there is a Hamiltonian cycle the total forced degree of each component must be even.

To obtain the results in the last column of Table 5.1, we first did the initial pruning, and then applied the test to the pruned graphs, using only the forced edges incident on degree two vertices.

## 6. Generalized Knight's Circuit Graphs

In this section we examine a graph class based upon the generalized knight's circuit problem in which the size of the knight's move is allowed to vary along with the size of the (rectangular) board. An instance of the generalized knight's circuit problem is a graph defined by the 4-tuple $(A, B) - n \times m$ where $A, B$ is the size of the knight's move and $n, m$ is the size of the board. The vertices of the graph correspond to the cells, and thus $|V| = nm$. Two vertices are connected by an edge if and only if it is possible to move from one vertex to the other by moving $A$ steps along one axis and $B$ along the other. (See Vandegriend, 1998 for more information about this problem.)

For this graph class there is no easy way to define phase transitions since there is no clear parameter which separates the Hamiltonian graphs from the non-Hamiltonian graphs (although Vandegriend, 1998 shows that there are ways of identifying groups of non-Hamiltonian graphs). Thus to find hard graphs, we look for graphs which take a significant amount of time to solve relative to their size. We perform 1 trial per graph (problem instance) and report the ratio of search nodes to number of vertices.

We examined a total of 300 generalized knight's circuit graphs over ranges of $A, B, n, m$ (Specific $A, B, n$ triplets with $m$ allowed to vary, for $A + B \leq 9$, $n \leq 13$, $m \leq 60$.) They ranged in size from 80 to 390 vertices. Of the 300 instances examined, 121 graphs (40 %) were found to be Hamiltonian and 141 graphs (47 %) were found to be non-Hamiltonian.

| search nodes | | # of trials | % of trials |
|---|---|---|---|
| $\leq$ | $2n$ | 1 | 0.8 |
| | $5n$ | 43 | 35.5 |
| | $10n$ | 37 | 30.6 |
| | $20n$ | 11 | 9.1 |
| | $50n$ | 8 | 6.6 |
| | $100n$ | 8 | 6.6 |
| | $200n$ | 2 | 1.7 |
| | $500n$ | 5 | 4.1 |
| | $1000n$ | 2 | 1.7 |
| | $2000n$ | 1 | 0.8 |
| | $5000n$ | 1 | 0.8 |
| | $10000n$ | 1 | 0.8 |
| | $20000n$ | 0 | 0.0 |
| | $50000n$ | 1 | 0.8 |

Table 7: Histogram of the search node ratio of our backtrack algorithm on 121 Hamiltonian generalized knight's circuit instances.

For the remaining 38 graphs (13 %) our backtrack algorithm failed (reached the 30 minute time limit), which implies these graphs are very hard for our backtrack algorithm.

A majority (91%) of the non-Hamiltonian graphs were solved without any search. However, a significant number of the remaining graphs took many search nodes to solve. 9 graphs (6.4%) took more than $10n$ nodes and 7 graphs (5.0%) took more than $100n$ nodes. The hardest graph took $\approx 11276n$ search nodes ($n = 324$). So while the majority of the non-Hamiltonian graphs were easy, a significant percentage of these generalized knight's circuit graphs were quite hard for our algorithm.

A larger variance in hardness was observed with the Hamiltonian graphs. Table 6 shows the distribution with respect to the number of search nodes required. Unlike $G_{n,m}$ and Degreebound graphs, these graphs could not be solved in only $n$ search nodes. Almost all the graphs required at least $2n$ search nodes. 33% of the graphs required at least $10n$ nodes, 11% required at least $100n$ nodes and the hardest graph required $\approx 34208n$ nodes ($n = 198$).

## 7. A Hard Constructed Graph Class

It is worthwhile when designing an algorithm to determine under what conditions and how frequently it might fail to perform and just how badly it might do. The measure can be in terms of how bad an approximation is, or how long an exact algorithm may take in the worst case. There is a long tradition of designing instance sets that foil specific combinatorial algorithms (Johnson, 1974; Mitchem, 1976; Olariu & Randall, 1989; Spinrad & Vijayan, 1985). Other special classes are intended to be more general, and are frequently based on certain features or constructs together with some randomization to hide the features (Culberson & Luo, 1996; Brockington & Culberson, 1996; Kask & Dechter, 1995; Bayardo Jr. &

Schrag, 1996). The $G_{n,m}$ class is frequently used to study graph algorithms over all possible graphs.

In this section we consider a special construction for a Hamiltonian graph which is extremely hard (exponential increase in difficulty with size) for our backtrack algorithm. It consists mostly of special constructs tied together with some randomly chosen edges. It bears some resemblance to graphs such as the Meredith graph (Bondy & Murty, 1976) used to disprove certain theoretical conjectures. This graph remains difficult when we vary the neighbour selection heuristic or pruning techniques used by our backtrack algorithm. The graph we construct we refer to as the Interconnected-Cutset ($ICCS$) graph.

Our class is intended merely to show that exponentially hard classes clearly exist for our algorithm, and many other backtrack algorithms using similar approaches. We do not claim our graphs are intrinsically hard, as there is a polynomial time algorithm that will solve this particular class.

The basic concept we use in constructing these graphs is the non-Hamiltonian edge, which we define as an edge which cannot be in any possible Hamiltonian cycle. Note that since the graphs are Hamiltonian, each vertex must be incident on at least two edges which are not non-Hamiltonian. Our goal is to force the algorithm to choose a non-Hamiltonian edge at some point. The key observation is that once such an edge is chosen, the algorithm must backtrack to fix that choice. With multiples of these bad choices, after backtracking to fix the most recent bad choice, the algorithm must eventually backtrack to an earlier point to fix a less recent bad choice, which means the more recent choice must be redone, with the algorithm making the bad choice again. The amount of work performed by the algorithm is at least exponential in the number of bad choices. See Vandegriend (1998) for more details.

The $ICCS$ graph is composed of $k$ identical subgraphs $ICCS_S$ arranged in a circle. To force the desired cycle we have a degree 2 vertex between each subgraph. Since each subgraph has a Hamiltonian path between the connecting vertices, the $ICCS$ graph is Hamiltonian. Due to the construction of the $ICCS$ subgraph, extra non-Hamiltonian edges can be added between different subgraphs. These edges help prevent components from forming during the search, which greatly reduces the effectiveness of the component checking search pruning. See Figure 5. Heavy lines are forced edges that must be in any Hamiltonian cycle.

Figure 6 contains a sample $ICCS$ subgraph. Non-Hamiltonian edges are denoted by dashed lines, and forced edges are denoted by heavy lines.

To see that the dashed lines cannot be part of any Hamiltonian cycle observe that any path through the $ICCS_S$ must enter and exit on an $S_C$ vertex, and between any two $S_C$ vertices in sequence the path can visit at most one $S_I$ vertex. Thus, each such path uses at least one more vertex from $S_C$ than from $S_I$. Since initially $|S_C| = |S_I| + 1$, any Hamiltonian cycle can enter and exit the $ICCS_S$ only once, and must alternate between $S_C$ and $S_I$ vertices. Since the $S_T$ vertices only have one edge leading to an $S_I$ vertex, these edges are forced. This also allows us to interconnect subgraphs without adding new Hamiltonian cycles by connecting vertices of $S_C$ of two different subgraphs (since these additional edges are all non-Hamiltonian edges). By interconnecting the subgraphs in this fashion, we strongly reduce the effectiveness of checking for components or cut-points during the search. In the current implementation, for each vertex in each $S_C$ we randomly choose a vertex in another $S_C$ and add the edge. Thus, the average number of such edges per vertex is a little less than two, since some edges may be repeated.

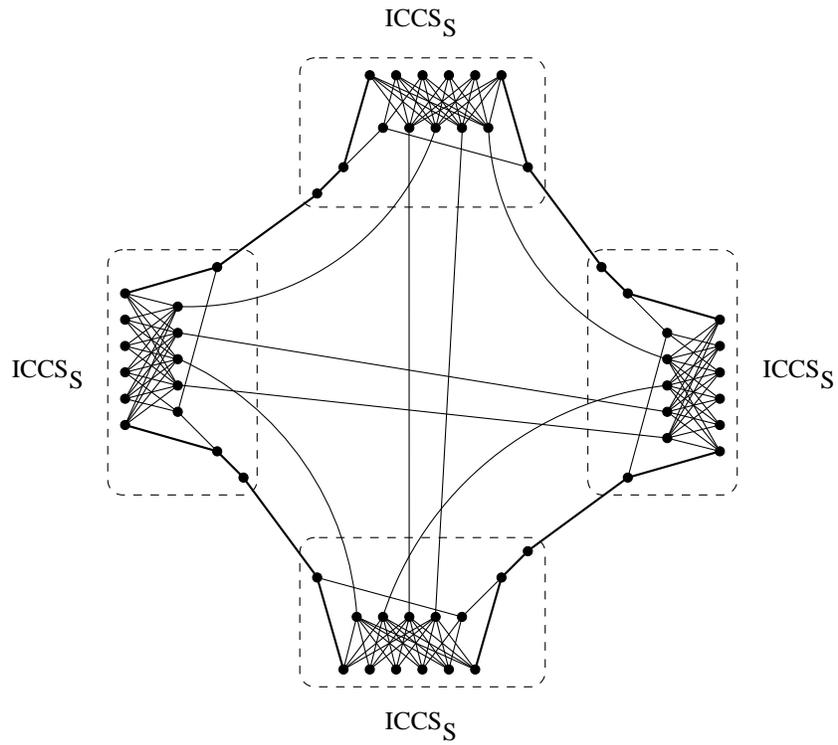

Figure 5: A sample *ICCS* graph.

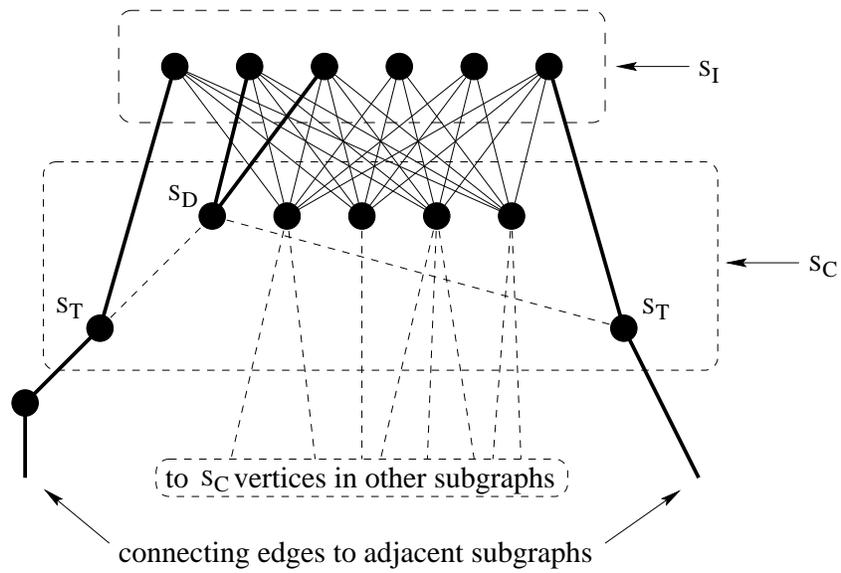

Figure 6: A sample *ICCS* subgraph $ICCS_S$.

One additional design element was added to handle various degree selection heuristics that our algorithm could use. At each stage in the search, the neighbours of the current endpoint of the partial path are arranged in a list to determine the order in which they will be chosen by our backtrack algorithm. There are 3 main heuristics: sorting the list to visit lower degree neighbours first, sorting to visit higher degree neighbours first, and visiting in random order. (Our backtrack algorithm normally uses the lower degree first heuristic.)

The $S_D$ vertex in the $ICCS$ subgraph is used to fool the low degree first heuristic. The $S_D$ vertex is only incident to the two $S_T$ vertices and to two vertices in $S_I$, which makes it degree 4. When the algorithm enters a subgraph from the degree 2 connecting vertex, it reaches one of the $S_T$ vertices. From the $S_T$ vertex, the choices are the $S_D$ vertex (degree 4) and the one $S_I$ vertex (degree $|S_C| - 2$, because it is not connected to the $S_D$ vertex and the other $S_T$ vertex). If $|S_C| > 6$ then the $S_D$ vertex will have a lower degree and thus will be chosen first.

The high degree first heuristic avoids following the edge from the $S_T$ vertex to the $S_D$ vertex, and instead goes to the $S_I$ vertex. From there it chooses one of the $S_C$ vertices (not including $S_D$ or the other $S_T$ vertex, which are not adjacent). From this point, its choice is one of the $S_I$ vertices (maximum degree $= |S_C| - 2$) or one of the $S_C$ vertices in a different subgraph (degree $\geq |S_C|$ if that subgraph has not yet been visited). Since the $S_C$ vertex normally will have a higher degree, the algorithm will follow the non-Hamiltonian edge to that vertex.

If the next neighbour is chosen at random, then from a $S_T$ vertex, the algorithm has a 50% chance of making the wrong choice. Similarly, at each $S_C$ vertex the algorithm has a small chance of following a non-Hamiltonian edge. As the number of subgraphs is increased, the probability of the algorithm making all the right choices rapidly approaches 0.

Another reason why the $ICCS$ subgraph is expected to be hard for a backtrack algorithm is that there are many possible paths between the two $S_T$ vertices. If a non-Hamiltonian edge has previously been chosen, then the backtrack algorithm will try all the different combinations of paths (and fail to form a Hamiltonian cycle) before it backtracks to the bad choice.

We performed experiments on various $ICCS$ graphs. We varied the number of subgraphs from 1 to 4, and varied the independent set size ($|S_I|$) from 6 to 8. We used our backtrack algorithm as specified in Section 3 with the addition of checking for components and cut-points during the search. We executed our algorithm 5 times per graph. Our results are listed in Table 7 for the low degree first heuristic. Our experiments using the other degree selection heuristics exhibited similar results.

We have also performed similar experiments using a randomized heuristic algorithm (Frieze, 1988; Pósa, 1976). Due to the significant difference in operation between this algorithm and backtrack algorithms, it easily solved these small $ICCS$ graphs. However its performance rapidly decreased as the graphs were increased in size.

The average degree of $ICCS$ graphs with more than one subgraph lies within the following range:

$$|S_I| - 2.5 + \frac{9.5}{|S_I| + 1} \leq \overline{d} \leq |S_I| - 2 + \frac{8}{|S_I| + 1}$$

| $n$ | $\#S$ | $|S_I|$ | Min | Median | Max |
|---|---|---|---|---|---|
| 14 | 1 | 6 | 14 | 14 | 210 |
| 28 | 2 | 6 | 606 | 616 | 3,777 |
| 42 | 3 | 6 | 10,467 | 47,328 | 112,795 |
| 56 | 4 | 6 | 6,538,842 | 32,578,160 | 36,300,827 |
| 16 | 1 | 7 | 16 | 48 | 112 |
| 32 | 2 | 7 | 13,056 | 21,797 | 70,949 |
| 48 | 3 | 7 | 1,350,084 | 5,247,287 | 8,027,520 |
| 18 | 1 | 8 | 18 | 54 | 270 |
| 36 | 2 | 8 | 283,164 | 430,620 | 750,211 |
| 54 | 3 | 8 | $> 1.2 \times 10^8$ | | |

Table 8: Search nodes required by our backtrack algorithm on *ICCS* graphs.

From this formula we see that as the size of each independent set is increased, the mean degree increases linearly. However, as the number of subgraphs is increased, the mean degree remains constant. The *ICCS* graphs remain hard over a very wide range of mean degrees (from $O(1)$ to $O(n)$). Therefore the average degree in this case is not a relevant parameter for determining hardness.

## 8. Conclusions and Future Work

Our backtrack Hamiltonian cycle algorithm found $G_{n,m}$ graphs easy to solve, along with a majority of Degreebound graphs. We have also performed similar experiments (Vandegriend, 1998) using a randomized heuristic algorithm (Frieze, 1988; Pósa, 1976) which had a high success rate on $G_{n,m}$ graphs, less so on Degreebound graphs. More interestingly, the existence of a phase transition for both problems did not clearly correspond to a high frequency of difficult instances. We suspect that other properties play a more important role than does the average degree. This is supported by our results on generalized knight's circuit graphs, which are all highly regular (with many symmetries), and for which the majority have average degrees between 4 and 8, compared to a mean degree $\leq 3$ on Degreebound graphs.

These results should not be surprising, since it has been shown that asymptotically for randomly generated graphs, when the edge is added that makes the last vertex degree 2, then with high probability the graph is Hamiltonian (Bollobás, 1984). In addition, efficient algorithms have been shown to solve these instances in polynomial time with high probability (Bollobás et al., 1987). Since vertices of degree less than 2 are a trivially detectable counter-indicator, it is hardly surprising that asymptotically determining Hamiltonicity of graphs in $G_{n,m}$ is easy.

We also observe that the performance of our backtrack algorithm can widely vary for a single graph due to the selection of the initial vertex. Multiple restarts of our backtrack algorithm after a time limit was reached often resulted in superior performance. We suggest a little randomization of the algorithm be used while empirically identifying intrinsically hard random instances of any problem.

## Acknowledgements

This research was supported by Natural Sciences and Engineering Research Council Grant No. OGP8053.